# Match Chat: Real Time Generative AI and Generative Computing for Tennis

Aaron Baughman
IBM
Cary NC, USA
baaron@us.ibm.com

Gozde Akay
IBM
Fredericton, Canada
Gozde@ibm.com

Eduardo Morales
IBM
Coral Gables FL, USA
Eduardo.Morales@ibm.com

Rahul Agarwal
IBM
New York NY, USA
rahul.agarwal@ibm.com

Preetika Srivastava
IBM
Houston, TX, USA
preetika.s@ibm.com

## ABSTRACT

We present Match Chat, a real-time, agent-driven assistant designed to enhance the tennis fan experience by delivering instant, accurate responses to match-related queries. Match Chat integrates Generative Artificial Intelligence (GenAI) with Generative Computing (GenComp) techniques to synthesize key insights during live tennis singles matches. The system debuted at the 2025 Wimbledon Championships and the 2025 US Open, where it provided about 1 million users with seamless access to streaming and static data through natural language queries. The architecture is grounded in an Agent-Oriented Architecture (AOA) combining rule engines, predictive models, and agents to pre-process and optimize user queries before passing them to GenAI components. The Match Chat system had an answer accuracy of 92.83% with an average response time of 6.25 seconds under loads of up to 120 requests per second (RPS). Over 96.08% of all queries were guided using interactive prompt design, contributing to a user experience that prioritized clarity, responsiveness, and minimal effort. The system was designed to mask architectural complexity, offering a frictionless and intuitive interface that required no onboarding or technical familiarity. Across both Grand Slam deployments, Match Chat maintained 100% uptime and supported nearly 1 million unique users, underscoring the scalability and reliability of the platform. This work introduces key design patterns for real-time, consumer-facing AI systems that emphasize speed, precision, and usability that highlights a practical path for deploying performant agentic systems in dynamic environments.

* Extended Support Staff = IBM Staffers who enabled the project listed in alphabetical order: Stephen Hammer (Technical Advisor), Hugo Hernandez (Cloud Eng), Kelly Hicks (Designer), Kevin Masters (Cloud Eng), Sara Perelman (Project Manager), Hannah Pippin (Annotator), Ravi Rajan (Cloud Eng), Karl Schaffer (Project Manager), Kevin Vaughan (Cloud Eng), Ryan Whitman (Cloud Eng)

## 1 Introduction

The concept of Generative Artificial Intelligence (GenAI) has undergone significant transformation alongside advances in AI research and computational modeling. The origins of GenAI trace back to the 1970's, rooted in symbolic and rule-based systems that relied on manually crafted logic and structured knowledge bases for reasoning tasks and decision-making [1,2,3,4,5]. These early AI systems typically employed case-based reasoning and deterministic rules, offering limited flexibility in learning or adaptation. During the 1980s, the field experienced a gradual paradigm shift toward neural network models, which introduced the ability to learn patterns directly from data [6]. This transition laid the groundwork for modern generative methods by enabling systems to infer representations and synthesize new outputs from training data. A breakthrough occurred in the mid 2010s with the introduction of Generative Adversarial Networks (GANs), which demonstrated that a generator and a discriminator network could engage in a cooperative training process to produce increasingly realistic synthetic content [7]. Subsequent progress was marked by the development of the transformer architecture and attention mechanisms, which revolutionized the way models handled sequence data and long-range dependencies [8]. These innovations directly enabled large-scale generative systems such as Generative Pre-trained Transformer (GPT-1), which exemplified the potential of pretraining followed by fine-tuning for language generation tasks [9]. In the 2020s, the field saw further advancements with the emergence of diffusion models, which offered high-quality generation capabilities across modalities such as image synthesis, audio, and video [10]. This technological evolution culminated in the widespread deployment of large-scale, multimodal GenAI systems, which now power applications in text generation, visual content creation, and conversational agents at scale [11,12].

As genAI continues to evolve and becomes increasingly embedded within enterprise systems, Generative Computing (GenComp) has emerged as a complementary paradigm that imposes structured, logic-based boundaries on these models. For example, IBM's application of GenComp frames Large Language Models (LLMs) not as opaque oracles, but as programmable infrastructure; components governed by the rigor of traditional software engineering practices [13]. This shift reflects a growing demand to incorporate deterministic control, modularity, and verifiability into generative systems that operate in production environments. Other sources [14, 15, 16, 17, 18] further describe

GenComp as an encompassing broad spectrum of computational techniques, including constraint optimization, rule-based systems, evolutionary algorithms, case-based reasoning, simulation engines, and logic inference. In these frameworks, logic-driven GenComp provides the structural support that guides or constrains generative outputs to ensure alignment with system-level goals, compliance, or user-defined parameters. When combined with GenAI, this hybrid approach leverages the creativity of probabilistic generation while ensuring outputs remain robust, interpretable, safe, and contextually grounded.

In this paper, we present the design, integration, and deployment of Match Chat, a real-time generative assistant developed for tennis fans. This system exemplifies the practical fusion of GenAI and GenComp through an Agent Oriented Architecture capable of answering natural language questions across 254 singles matches. We propose architectural strategies for combining large-scale pretrained models with logic-driven rules and algorithmic controls to enable agentic behavior in a live, user-facing setting. Our key contributions include:

- A large-scale, real-time assistant that combines GenAI with GenComp.
- A scalable AOA framework that supports GenAI concurrency and model optimization.
- A GenComp-guided user experience that promotes interpretability and seamless interaction.
- GenAI caching and shielding mechanisms that enhance reliability at scale.

## 2 Related Works

The development of generative AI systems capable of operating in real-time across multiple domains has garnered growing attention in recent years. Foundational resources such as the Schema-Guided Dialogue (SGD) dataset have provided critical infrastructure for building scalable, task-oriented assistants, offering over 16,000 multi-domain conversations for training and evaluation [20]. Additional datasets such as DialogStudio, MultiWOZ, cPAPERS, U-NEED, and GenQA have expanded coverage across domains, including finance, health, sports, and entertainment [21–25].

Despite progress across these verticals, large-scale real-time GenAI and GenComp assistants remain limited in practical deployment. A notable example includes Amazon Pharmacy, where a consumer-facing assistant facilitates prescription processing and healthcare queries through a structured, real-time workflow [30].

Our work extends this emerging field with Match Chat, a real-time, consumer-facing assistant that delivers immediate access to both raw and derived tennis data metrics and insights. Match Chat supports real-time interaction synchronized with live match activity, exemplifying a scalable model for multi-domain, high-frequency generative applications.

### 2.1 Real Time Generative AI Systems

Recent advancements in GenAI have enabled real-time deployment of interactive systems but achieving low-latency response at scale remains a fundamental challenge. Several methods have been proposed to address inference latency and computing resource consumption, including prompt caching, shallow reranking, speculative decoding, State Space Model (SSM), and model distillation. For example, Prompt Cache is a method for reusing previously computed attention segments that builds on Key-Value Caching [31,32]. Further, speculative decoding allows for tokens to be generated in parallel by fast models and then verified by a larger model to reduce decoding time [33]. Another technique, called model distillation, contributes to latency reduction by training smaller models on knowledge extraction from larger models, which tends to contribute towards more performant models [34]. Alternative architectures, such as State Space Models (SSMs) like Mamba, uses a sequence of equations to model sequential data, offering a faster and more memory-efficient context generation approach than transformers [35,36]. Another class of improvements are on-computing device improvements. For example, many cloud providers are utilizing quantized models such as Meta's Llama Nano series, Google's Gemini Nano, and gpt-oss-120b to create more efficient models [37,38,39].

Even though commercial applications of GenAI are becoming more common, latency, accuracy, evaluation, and cost of systems are a primary concern. A Forbes article mentioned that optimizing for one of cost, latency, or relevance is often at the expense of another [40]. As enterprise systems expand to support more users and larger or a plurality of models, the resources for training and inference are prohibitive in footprint and cost [41]. As models are scaled out and real-time latency grows, the results can become less predictable [42]. As a result, real-time and large-scale consumer-facing GenAI remains an open problem.

### 2.2 Generative AI with Predictive Modeling

The combination of GenAI with predictive modeling creates new ways of deriving statistical insights for scene interpretation across different domains. One approach uses conditional generative models to change the results of a predictive model by minimizing a loss function comprised of label changes, GenAI sparsity creation, and out of distribution penalties [43]. This work's goal is to alter a predictive model's output under altered conditions [43]. In a different direction where outputs of a predictive model are input into a GenAI model, Felice's work depicts how insights were communicated at the 2024 Olympic handball tournament [44]. GenAI translated the predictive insights into human-understandable explanations that highlighted the reasons for a particular prediction [44]. In another example, conditional GANs for financial market simulation was combined with classical time-series forecasting models to jointly predict changes in financial markets [45].

Extending to full scene narration, GenAI has been utilized to describe sports and music scenes using raw data and insights by predictive models [46]. This large-scale implementation of narrative generation provides an example of how the contribution of different features can contribute to rich storytelling [46]. In addition, a knowledge graph implementation uses a predictive model to identify key events in live broadcasts and then fuses that together with Convolutional Neural Networks (CNN) and a transformer encoder [47]. These methods reflect a broader shift where traditional Machine Learning (ML) and GenAI complement each other through conditioning in both directions.

## 2.3 Agent Oriented Architecture

Service-Oriented Architecture (SOA) enables the modularization of functionality into discrete, self-contained services [48]. Agent-Oriented Architecture (AOA) applies similar modular principles, segmenting sensing, reasoning, and acting into structured components that support goal-directed behavior. The Belief-Desire-Intention (BDI) model illustrates how agents sustain internal state through beliefs, pursue motivational goals, and execute plans that support responsive and adaptive behavior [49]. Foundational agent communication and coordination capabilities are supported by protocols such as the Model Context Protocol (MCP), Agent Communication Protocol (ACP), and Agent-to-Agent Protocol (A2A), which facilitate collaborative interactions among agents and tools [50,51,52]. Together, these elements establish a framework for constructing multi-agent, goal-driven systems in which GenAI models can break down objectives, coordinate across subtasks, and adjust to changing contexts.

Frameworks like ReAct unify reasoning and action, allowing agents to combine logical inference with external tool execution [51]. Systems developed with ReAct have demonstrated performance gains on tasks such as HotpotQA and ALFWorld [51]. Extensions such as AutoGPT and BabyAGI build upon ReAct by integrating memory-enabled planning into recursive loops, enabling complex multi-step workflows [52]. Tools like LangChain and LangGraph further enhance agent orchestration by offering configurable pipelines and managing stateful interactions between components [53]. Specifically, LangGraph encodes agents in a graph-based structure, where nodes are agents and edges represent communication between agents [54].

## 2.4 GenAI Assistants

Assistants and agents are often used interchangeably, but they differ in terms of their functionality and autonomy. An assistant is typically a reactive system that responds only when prompted by a user [55]. In contrast, an agent is proactive, goal-driven, and capable of interacting with tools and external environments [55]. In our work, Match Chat functions as an assistant that is backed by a network of agents. More broadly, assistants and agents can be layered where the assistant acts as the user interface, while agents manage more logically complex operations behind the scenes.

The public-facing use of GenAI assistants, such as ChatGPT, Gemini, Claude, and Perplexity, has grown rapidly with a combined global user base estimated at 987 million people, roughly 12% of the world's population [56]. Other sources suggest that 800 million to 1 billion users access ChatGPT weekly, contributing to around 2.5 billion prompts daily [57]. Each day, between 115 to 180 million users actively engage with GenAI, which intersects with assistants [58]. Between 2024 and 2025, the adoption of GenAI assistants has increased by a factor of 4, indicating their increasing integration into daily life [59].

ChatGPT has set the standard and benchmark for modern conversational GenAI assistants by leveraging OpenAI's Generative Pre-trained Transformer (GPT) family, such as GPT-3.5, GPT-4, GPT-4.5, and GPT-5, depending on the subscription tiers [60,61]. In August 2025, OpenAI released open weight models, gpt-oss-20b and gpt-oss-120b, allowing developers to create GenAI applications and assistants such as deep browsing, python tool use, and developer functions [62]. Another popular GenAI assistant is Google's generative AI in search [63]. Although limited technical details are available, it is likely based on techniques developed from PaLM and/or Gemini [64,65]. As a broader trend, many GenAI assistants are being adapted or fine-tuned for specific application domains while still leveraging generalist architectures.

Domain-adapted assistants leverage targeted adaptation strategies such as fine tuning, in-context learning, Reinforcement Learning with Human Feedback (RLHF), Group Relative Policy Optimization (GRPO), mixture of experts, etc., to close domain-specific performance gaps while maintaining generalist capabilities [66,67,68]. For example, in finance, Klarna's AI chatbot, developed with OpenAI, served over 150 million users in 2024, lowering answer time from 11 minutes to under 2.5 minutes [69]. In healthcare, a Conversational Patient Assistance Triage (C-PATH) system was fine-tuned on medical data to support symptom recognition and department routing for patients [70]. Within the multimodal domain, a Vision-Language Model adapted for soccer understanding enabled an improvement of visual question-answering performance [71]. Building on these advances, our system extends prior work by enabling high-volume, real-time question-answering, delivering live statistical insights during sports matches.

## 3 Live Event Information

In 2025, over 1.5 million fans attended the US Open and Wimbledon in person, while an additional 30 million engaged remotely via mobile and desktop platforms. With 127 matches per main singles draw for both men and women, totaling 508 matches, spread across 35 courts, 18 at Wimbledon and 17 at the US Open, the volume and pace of the tournament action presented a significant information challenge. For instance, the Arthur Ashe Stadium alone can host over 23,700 spectators, contributing to a highly dynamic and distributed event environment.

To address this, we introduced the Match Chat assistant, powered by AOA, as a centralized interface for fans both onsite and remote to obtain real-time insights and answers to their questions about live matches. The assistant leveraged in-context learning from a rich dataset that included player profiles, match logistics, live statistics, and hundreds of performance metrics. By making this level of data instantly accessible before, during, and after matches, Match Chat enabled fans to spend less time searching for information and more time engaging with content. This led to a double-digit increase in average user session duration, along with comparable growth in unique page views across both tournaments. Ultimately, the assistant provided an engaging, informative, and scalable experience, helping fans follow the narrative of each match through a factual and approachable conversational interface, all made possible by the underlying diversity of data described in the following section.

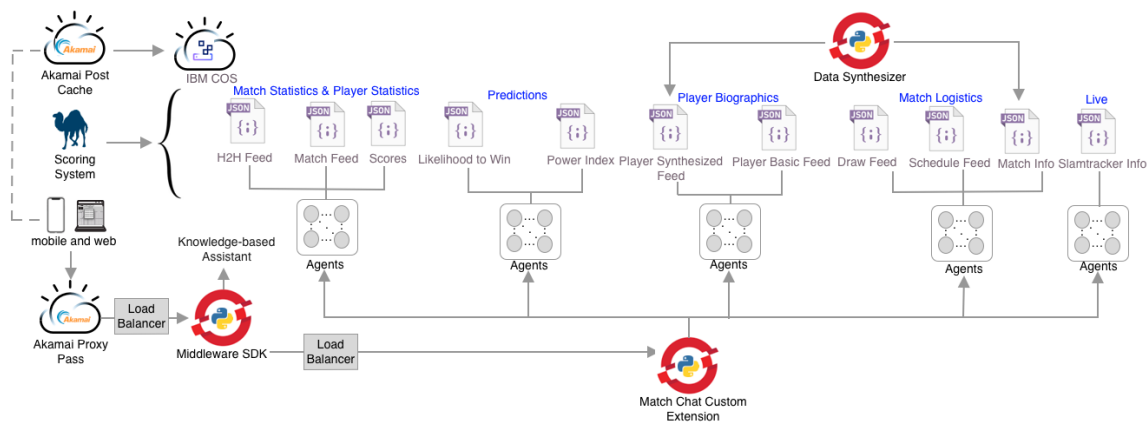

Figure 1: Match Chat Architecture deployed at Wimbledon & US Open 2025

## 3.1 Match Data

The tennis match data layer forms the foundation of Match Chat's ability to deliver timely, relevant, and context-aware insights during live Grand Slam matches. Multiple streams of structured data are ingested in real time and embedded into text-rich prompts, allowing the system to understand and communicate complex match dynamics. As play unfolds, over 300 statistical measures are accumulated per match, including core performance indicators such as total aces, first serve attempts and success rates, second serve percentages, break points won, and unforced errors. These statistics are calculated for each player individually and update continuously within seconds of the live event.

Beyond basic metrics, the data architecture captures serve, return, rally, directional patterns, and distance run, enriching the understanding of point construction and player behavior. Tiebreaks, set and game scores, as well as point-by-point durations, are recorded with high temporal resolution. A dedicated point-by-point feed narrates the match by logging every point, including details such as break points, unforced errors, serve speeds, and rally lengths. This feed is segmented by set and game within Match Chat, allowing fans to explore specific match moments in detail.

In addition to performance data, logistical information such as match state, court assignment, schedule, and draw position are integrated to provide situational awareness. Temporal data like prior and upcoming matches for each player offer context around a player's draw difficulty. Further enrichment is provided through court-level metadata, including seating capacity, surface type, location, environmental factors, and historical play patterns.

This layered and temporally synchronized dataset transforms the traditionally overwhelming stream of match data into an accessible and conversational experience. By organizing and presenting match data in a contextually relevant way, Match Chat allows users to navigate the statistical narrative of the match effortlessly, setting the stage for deeper understanding about the players described in the following section.

## 3.2 Player Data

Prior to the start of each Grand Slam tournament, certain player-related data is held static to reflect the state of play at that point in time. For example, historical head-to-head records between two players are determined before the tournament begins and remain constant throughout its duration. The data feed includes aggregated match win counts along with associated metadata such as the match location and date. Core biographical attributes such as tournament seed, tour ranking, nationality, gender, height, weight, and date of birth, etc., are also held constant for the entirety of the tournament. In addition to the static biographical information, ancillary player data includes curated genAI summaries and factual information, which offer expanded insights into the players' career and recent performance trends. These facts provide a richer context for understanding a player's competitive background.

Another category of player data consists of historical performance statistics that remain constant for the duration of the tournament. These include cumulative career win–loss records, match statistics for the current year up to the start of the event, and Grand Slam-specific achievements such as total titles at the Australian Open, Roland Garros, Wimbledon, and the US Open. Historical comparisons, such as a player's result in the current Grand Slam during the previous year are also included to enable longitudinal analysis. This collection of descriptive and comparative statistics supports informed, in-depth user queries via the Match Chat system.

However, some data elements are dynamically updated as the tournament progresses. For example, total court time is accumulated after each match. Additionally, in-match player performance statistics are captured within the match-specific data feeds described in Section 3.1.

## 3.3 Predictions

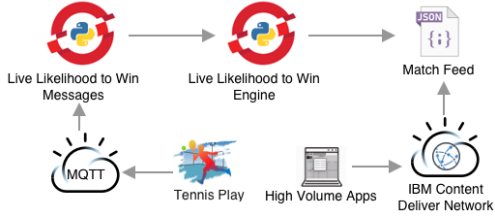

Figure 2: Live Likelihood to Win

Match Chat supports two types of predictive models for professional tennis: a pre-match likelihood to win and a live (in-play) likelihood to win, each computed separately for men's and women's singles matches. The Pre-Match Likelihood to Win estimates the probability that player $x$ will win a match before it begins, based on static pre-match evidence. When no prior head-to-head history exists between the two players, the probability $P_{static,nh}(p_x = wins, p_y = loses|e_p)$ is estimated using an XGBoost decision tree model trained on historical match data shown in Equation 1. Here, $e_{nh}$ represents a set of predictors including player age, the proprietary Watson Power Index, recent form, surface preference, and historical win ratios.

$$P_{static,nh}(p_x = wins, p_y = loses|e_p) = dt_{nh}(e_{nh}) \quad (1)$$

If players have a documented head-to-head record, the model incorporates additional features from this history. In that case, the conditional probability $P_{static,h}(p_x = wins, p_y = loses|e_p)$ is modeled similarly by Equation 2, where $dt_h$ denotes a distinct XGBoost model and $e_h$ extends the feature set $e_{nh}$ to include head-to-head metrics such as number of matches, sets won, and game ratios.

$$P_{static,h}(p_x = wins, p_y = loses|e_p) = dt_h(e_h) \quad (2)$$

As a match progresses, the system transitions to a point-by-point live prediction model. The Live Likelihood to Win, denoted $P_{live,point}$ is conditioned on the current game state $e_{live,p}$, which includes the current score (points, games, sets), as well as dynamic features such as momentum, fatigue decays, and advantage factors.

$$P_{slive,point}(p_x = wins, p_y = loses|e_{live,p}) \quad (3)$$

The evolving match state at any time $t$, represented as $S(t)$, is defined in Equation 4 as a tuple of sets, games, and points for each player. A transition function $f(S(t))$ in Equation 5 updates the state when a player wins or loses a point. This function encapsulates official tennis rules, such as four points win a game (by two), six games win a set (by two), tiebreak conditions, and match formats (best-of-three or five sets).

$$S(t) = (sets_{p1}, sets_{p2}, games_{p1}, games_{p2}, points_{p1}, points_{p2}) \quad (4)$$
$$S(t+1) = f(S(t)) \quad (5)$$

At each point, the system calculates momentum for each player, denoted $M_x(t)$ which reflects performance dynamics relative to baseline pre-match expectations. Momentum is updated incrementally based on the outcome of each point. For a point won by player $x$, the momentum delta $\Delta_x(t)$ is computed as a positive value weighted by a context-aware function $g(S(t))$, emphasizing high-pressure scenarios like break points as seen in Equation 6. Conversely and in Equation 7, if the point is lost, momentum is reduced accordingly

$$\Delta_x(t) = \alpha \cdot g(S(t)) \quad (6)$$
$$\Delta_x(t) = -\beta \cdot g(S(t)) \quad (7)$$

Momentum is updated recursively by Equation 8 and decayed over time to prioritize recent performance. The decayed momentum $M_x^{decay}(t)$ is calculated using an exponential decay factor $\lambda$ and match completion percentage $c(t)$ like Equation 9.

$$M_x(t) = M_x(t-1) + \Delta_x(t) \quad (8)$$
$$M_x^{decay}(t) = M_x(t) \cdot e^{-\lambda \cdot c(t)} \quad (9)$$

The scaled momentum $M_x^{scale}$, representing the influence of player $x$'s momentum relative to both players, is then computed by Equation 10.

$$M_x^{scale}(t) = \frac{M_x^{decay}}{M_x^{decay} + M_y^{decay}} \quad (10)$$

The final live likelihood to win $P_{live}(t)$ is calculated as a weighted combination of the static pre-match probability and the scaled, decayed momentum. The weight $w(t)$ depends on the number of remaining points needed by each player, ensuring the probability appropriately reflects closeness to victory like Equation 11.

$$P_{live}(t) = P_{static} \cdot (1 - w(t)) + M_x^{scale}(t) \cdot w(t) \quad (11)$$

Additionally, a set booster is applied to emphasize the number of completed sets already won by a player. In Equation 12, this final adjusted probability $P_{live}^{boost}(t)$ is scaled based on the number of sets won by player $x$ where $s_x$ is the number of sets won and $S$ is the total sets required to win the match.

$$P_{live}^{boost}(t) = P_{live}(t) \cdot (1 + \frac{s_x}{S}) \quad (12)$$

The full prediction engine architecture is depicted in Figure 2. Real-time match data including point outcomes, scores, and player statistics is streamed via an MQTT-based publish-subscribe system. The messages application subscribes to relevant topics and ingests incoming data to update the match state and recalculate player probabilities on each point. The engine applies the previously mentioned transition functions, momentum updates, decay and scaling factors, and final set boosters to compute the evolving win probabilities. The output is structured as a JSON Live Likelihood to Win Match Feed, which is published through a CDN for scalable, low-latency distribution. This design ensures reliable and high-performance delivery of live win probabilities to millions of concurrent users across tournament websites, mobile applications, and third-party integrations keeping latency below the time taken for 2 consecutive points in a match, providing real-time predictions.

## 3.4 Knowledge Base

For queries that fall outside the domains of players, tennis play, or match-related data, Match Chat leverages a supplemental knowledge base. When the classification model described in Section 5.3 is unable to categorize a user query into predefined classes such as player statistics, predictions, biographical data, logistical information, live updates, or match statistics, the query is redirected to the knowledge base. In addition, if the agentic system outlined in Section 5.4 encounters a knowledge gap and cannot answer the query, the request is routed to this knowledge source.

These two routing mechanisms extend Match Chat's capabilities to address a broader range of user questions, including those but not limited to food vendors, parking, ticketing, venue logistics, and weather conditions. The implementation of the knowledge base varies across Grand Slam tournaments to reflect their differing application and operational needs.

At Wimbledon, the system linked to IBM Watson Assistant to host rule-based question routing to specific actions. This approach utilized question embeddings and similarity metrics to determine the most relevant data source or response. Responses were based on the latest tournament information maintained on official sites. In contrast, for the US Open, Match Chat integrated with an external customer care assistant developed by Satisfi Labs [72]. This deep-linking approach enabled seamless handoff between domain-specific assistants, allowing each system to focus on its area of expertise while maintaining continuity in the user experience.

## 4 User Experience

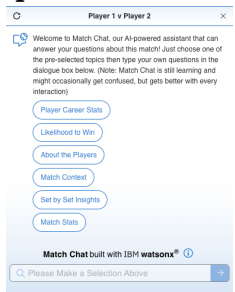

Figure 3: Match Chat user experience

The Match Chat user experience was intentionally designed around the principle of affordance and intuitiveness, ensuring that users can readily perceive how to interact with the system based solely on its visual and functional design. The goal was to create a frictionless, user-friendly interface that accommodates both guided and open-ended interactions. The system is accessible through two primary entry points. First, users navigating from a live match score page can launch Match Chat either by selecting a predefined question or by submitting a free-form query. Second, Match Chat can be invoked from other assistant systems, allowing users to initiate their own queries directly. In both scenarios, the system either delivers an immediate answer to a predefined query or initiates a dialog aimed at guiding the user toward a more precise line of questioning.

Upon activation, users are presented with a set of high-level tennis-related categories such as Player Career Stats, Likelihood to Win, About the Players, Match Content, Set-by-Set Insights, and Match Statistics as displayed in Figure 3. After selecting a category, users may either choose from a list of optional sub-questions or submit a free-text query. This tiered interaction model provides an approachable interface that requires no prior onboarding or technical knowledge. When a user submits a custom query, the classification model described in Section 5.3 evaluates whether the question aligns with the initially selected category. If misalignment is detected, the assistant prompts the user to confirm or adjust the focus area. This mechanism gives users control over the level of specificity they provide, whether through structured interface selections or natural language input.

Explicit user selections such as choosing Set-by-Set Insights and specifying a particular set and associated statistics provide semantic cues that streamline data retrieval and routing within the agentic system outlined in Section 5.4. These cues reduce the computational complexity otherwise required to resolve natural language queries. Otherwise, when users submit open-ended questions, additional processing is required by the generative AI components to classify and extract relevant information. Regardless of whether a user engages through structured inputs or fully open-ended questions, Match Chat is architected to deliver responsive and contextually accurate outputs. The layered user interface design masks the underlying architectural complexity, allowing users to benefit from advanced AI capabilities without friction or delay.

## 5 Match Chat Architecture

The Match Chat system, $MC$, was architected to balance generative accuracy and response speed by integrating both GenAI and GenComp components, as illustrated in Figure 1 and equation 13. Central to this architecture is the Data Synthesizer, $S_i$, a content generation module responsible for producing static information on players, courts, and other foundational tennis entities. This synthesized content, $D_o$, is used both in direct user-facing responses and as contextual augmentation for prompts to LLM's. The content is continuously streamed into an AI content hub, where human editors can optionally review or revise material prior to publication. Once validated, the static data is injected into conversations with sub-second latency, ensuring rapid user interactions. Other components that makeup Match Chat include a set of agents, $A_j$, HAP pipeline, $H_l$, question classifier, $C_m$, and a set of data feeds, $F_n$.

$$MC = \{S_i, A_j, K_k, H_l, C_m, F_n, D_o\} \quad (13)$$

In parallel to $MC$, real-time match data is processed through a publish/subscribe messaging model using MQTT topics. When data events such as match state changes, statistical updates, or player-related metrics are published, subscribing services parse and transform this information into structured JSON feeds, $L2W$, for downstream consumption. Additional feeds that are used by Match Chat are shown in Equation 14 such as Head-to-Head statistics $H2H$, Match Statistics $MStats$, Tennis Scores $Sc$, Likelihood to Win $L2W$, Watson Power Index $WPI$, Player Statistics $PStats$, order of play $Draw$, Match Logistics $MLog$, and point by point information on Slam Tracker $Slam$.

$$F_n = \{H2H, MStats, Sc, L2W, WPI, PStats, Draw, MLog, Slam\} \quad (14)$$

Incoming user traffic from web and mobile platforms, $x_i$ from user $i$, is first routed through an Akamai caching layer, configured with a Time-to-Live (TTL) of 2 seconds. This short TTL allows rapid response to repeated, identical requests, protecting backend systems from sudden traffic surges. Static content, including player biographies and match schedules, is also delivered via Akamai CDN, backed by IBM Cloud Object Storage as the origin server.

Requests that pass through the caching layer are handled first by the Middleware Application, $MW$, a horizontally scaled service deployed across three regions with two in AWS and one in IBM Cloud. Each of its 30 replicas per region operates with 2 CPUs, 4 GB RAM, and 500 GB of shared disk. The Middleware Application performs the initial request analysis, including user intent routing, category prediction, and input normalization. If the user has specified a particular tennis category or subcategory, a lemma standardization process selects a canonical form of the question. In these cases, an Answer Synthesizer attempts to respond without directly invoking a foundation model, which is part of a broader GenAI shielding strategy described in Section 5.1. If no predefined category is identified the system dynamically classifies the user query with $C_m$. When classification confidence is low or if the inferred category deviates from the user's intent, the system seeks explicit user confirmation. Queries deemed out-of-scope are routed to a knowledge base, $K_k$, retrieval workflow that is detailed in Section 3.4.

$$y_{i,1} = MW(x_i, S_i, K_k, H_l, C_m, F_n, D_o) \quad (15)$$

All user queries are also passed through a Hate, Abuse, and Profanity (HAP) detection pipeline, $H_l$, as described in Section 5.2. If inappropriate language is detected, the system halts processing and sends an instructive message to the user. For queries that pass all validations and remain in scope, they produce output $y_{i,1}$ for user $i$, which is passed to the Match Chat Custom Extension Application, $CE$, for further processing.

$$y_{i,2} = CE(y_{i,1}, S_i, A_j, F_n, D_o) \quad (16)$$

The Custom Extension Application, $CE$, is deployed with 60 replicas per region across two AWS regions and one IBM Cloud region. Each instance operates with 4 CPUs, 12 GB RAM, and access to a 500 GB shared mount where relevant AI models and assets are stored. This application hosts a bank of LangGraph agents, $A_j$, each representing a modular, agentic graph structure. Upon receiving a user request, the application selects the appropriate LangGraph and executes a tool within the graph that queries the real-time streaming JSON feeds, $F_n$, and content that has been reviewed by human annotators, $D_o$. The resulting data is embedded into a prompt and forwarded to a LLM, enabling rich, context-sensitive responses. Additional details on the agentic framework and LangGraph tooling are provided in Section 5.4. The resulting output, $y_{i,2}$ for user $i$ has been corrected and edited by a set of agents within $A_j$. The output of $CE$ is post processed, $POST$, within $MC$ that is presented to the user as $y_i$.

$$y_i = POST\left(CE\big(MW(x_i, S_i, K_k, H_l, C_m, F_n, D_o), A_j\big)\right) \quad (17)$$

Finally, traffic to both the Middleware $MW$ and Custom Extension $CE$ Applications is managed by an ingress load balancer, while outbound traffic from both services is distributed through a software-based load balancer. For example, during Wimbledon, the knowledge base was deployed in London and Dallas regions with traffic routing decisions based on regional performance metrics. Outbound responses from the Custom Extension Application were balanced across Frankfurt, Dallas, London, and Toronto, optimizing latency and throughput for global users.

## 5.1 Generative AI Shielding

Scalability remains a primary challenge in real-time, consumer-facing genAI systems such as Match Chat. A critical bottleneck emerges from the memory constraints required to host large models within GPU memory, especially under high user traffic. Based on prior usage patterns from major tennis Grand Slam events, Match Chat must accommodate peak loads of up to 900 RPS and sustain 450 RPS. These operational demands significantly exceed the standardized throughput limits of cloud-based GenAI platforms. For example, Amazon Bedrock and Azure OpenAI typically support only around 17 RPS under default configurations [74, 75]. This disparity necessitated architectural innovations that reduce dependency on LLM inference.

When sufficient context is provided by the user to infer a reliable response, the middleware system, $MW$, delegates the request to a lightweight data synthesizer, $S_i$. This component avoids invoking LLMs by using 135 pre-defined data synthesis patterns, which retrieve structured data from live feeds $F_n$ and construct coherent natural language responses. Additionally, for queries related to static attributes such as player biographies, the classifier $C_m$ routes the request to a GenAI-based static knowledge store $D_o$, allowing for immediate and deterministic responses. Collectively, this shielding approach offloads more than 50% of user traffic from GenAI inference, preserving compute resources and decreasing latency.

In cases where GenAI is necessary, the user query is routed from $MW$ to $CE$, which invokes the agentic graph $A_j$ as described in Section 5.4. To safeguard the user experience, LLM requests are wrapped in timeout constraints. If a response is not returned within a user study defined 6 seconds, a fallback mechanism activates a heavyweight version of the data synthesizer, also denoted $S_i$. Unlike its lightweight counterpart, the heavy $S_i$ incorporates an embedding-based matching layer, using the MiniLM-L6-v2 model to compute the cosine similarity between the input query and a corpus of 600 relevant generated sentences. The sentence with the highest semantic alignment is selected and returned as a fluent response. This multi-tiered shielding system provides robustness and responsiveness at scale, even under strict throughput and latency constraints.

## 5.2 Hate, Abuse, and Profanity Pipeline (HAP)

A key module in the $MW$ application is the HAP pipeline, denoted as $H_l$. This pipeline is composed of a structured sequence of processing engines $E_{l,p}$, where each component serves to enhance the trustworthiness, accuracy, and safety of natural language inputs. The pipeline consists of five primary stages:

1. Proper Noun Disambiguation ($E_{l,1}$): This module ensures that mentioned player names are correctly spelled and contextually relevant to the selected tennis match. A distilled model was utilized to compute word parts of speech.
2. Profanity Detection ($E_{l,2}$): Utilizes standard profanity libraries, such as python's profanity, to flag explicit language.
3. Custom Derogatory or Slur Language Detection ($E_{l,3}$): Implements a curated list of both commonly known and domain-specific derogatory expressions that may target individual players.
4. Suspicious Pattern Detection ($E_{l,4}$): Detects adversarial inputs such as prompt injections or code injection

   attempts by analyzing input patterns for malicious or suspicious behavior.
5. Pronoun Correction and Disambiguation ($E_{l,5}$): Ensures that natural language gender aligns with the selected player's profile. For instance, ATP players are referred to using masculine pronouns and WTA players with feminine pronouns.

As shown in Equation 18, these five engines work together to preprocess and sanitize user inputs while maintaining high throughput and low latency, which is essential for real-time applications.

$$H_l = \{E_{l,1}, E_{l,2}, E_{l,3}, E_{l,4}, E_{l,5}\} \quad (18)$$

To support efficient deployment and inference, the HAP pipeline integrates a distilled POS tagging model. The original teacher model, TweebankNLP/bertweet-tb2_ewt_pos-tagging, is based on a RoBERTa-base architecture with approximately 110 to 135 million parameters and a disk size of 537 MB. To reduce memory and computational overhead, the model was distilled into a student model based on distilbert-base-uncased, reducing size to 250 MB and parameter count to 66 million. Training was conducted on the Universal Dependencies English Web Treebank dataset, which contains 12,543 training, 2,000 validation, and 2,077 test samples. After three epochs of training, the distilled model achieved:

| Recall | Precision | Accuracy |
|--------|-----------|----------|
| 93.7%  | 93.3%     | 94.7%    |

Table 1: Resulting distilled model performance.

The distillation process resulted in a 53.4% reduction in disk size and over 40% fewer parameters, with minimal degradation in performance, making it suitable for Match Chat. The training hyperparameters, such as a 2e-5 learning rate, 3 epochs, and 0.01 weight decay, were adopted based on best practices for BERT-style architectures [73].

## 5.3 Predictive Modeling

The GenComp paradigm was employed to route user queries into the appropriate tennis category, enabling downstream agentic components to operate on targeted data subsets rather than extracting from the full set of data feeds $F_n$. Accurate query categorization also facilitated multi-turn interactions with the user, allowing the system to incrementally refine the information required to fulfill the query. The first step in this process involved transforming the user's natural language question into a structured representation suitable for classification.

First, a keyword detection module extracted class-specific lexical cues. When domain-relevant terms were identified, binary features were set to indicate the presence of these keywords. In parallel, the input sentence was encoded into a dense vector using the all-MiniLM-L6-v2 sentence embedding model. This embedding model, which consists of approximately 22 million parameters and occupies around 100 MB of disk space, was selected due to its balance of computational efficiency and semantic representation capability, which is an ideal fit for high-throughput classification in production environments.

A test and train dataset consisting of 627 manually annotated user questions and augmented with WordNet for a total of 1379 exemplars was embedded using this model. These embeddings were used to train a Random Forest classifier comprising 100 decision trees, with a cross-entropy loss used to evaluate classification performance. The target classes included: match statistics, player statistics, outcome predictions, player biography, match logistics, and live point-by-point data.

To handle cases of low classification confidence, the model's output probabilities were aggregated and compared against a pre-defined z-score threshold of 1.2. If no class surpassed this threshold, the query was redirected to the Knowledge Base module described in Section 3.4. After training, the classifier achieved a macro-average precision of 84%, with its Receiver Operating Characteristic (ROC) curve depicted in Figure 5.

Both the sentence embedding model and the trained Random Forest were deployed to a shared mounted disk, accessible by all regional replicas of the system. This architecture allowed the classifier to execute locally with the $MW$ layer, ensuring low-latency prediction and high availability across global deployments.

## 5.4 Performant Agents

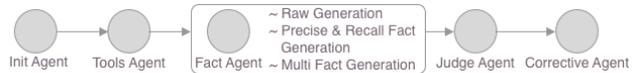

Figure 4: Match Chat Agentic Architecture

The Match Chat system, $MC$, operates as a collaborative framework supported by a sequence of specialized agents, denoted $A_j$, embedded within a directed graph structure $G$. Each vertex in the graph represents an individual agent, as defined in Equation 20.

$$G = (V, E) \quad (19)$$
$$V = A_j = \{A_1, A_2, A_3, A_4, A_5\} \quad (20)$$

Edges between the vertices shown Equation 21 define the communication and processing flow between agents.

$$E \subseteq V \times V \quad (21)$$

This multi-agent architecture enables a pipeline for generating fluent and contextually relevant responses to natural language queries. Each agent maintains its own specialized logic that contributes to the eventual response to the user.

1. Initialization Agent, $A_1$
   The first agent initializes a global data structure that maintains shared state across subsequent agent invocations. This ensures continuity and data persistence throughout the interaction.
2. Tools Agent, $A_2$
   Agent $A_2$ identifies and selects an appropriate data extractor based on the query's tennis-related category such as match statistics. This selection governs downstream data retrieval for response generation.
3. Fact Agent, $A_3$
   Agent $A_3$ performs multiple concurrent operations to generate candidate responses. It prioritizes outputs based on a hierarchy of fluency and latency. The preferred method is to pass the tool-selected output from $A_2$ directly into an LLM, specifically LLaMA 3-3 70B Instruct, selected due to its quantization representation on IBM-hosted hardware and prior benchmarking performance [46]. At this stage, Agent $A_3$ extracts the relevant key–value pairs from JSON, composes a brief explanatory context highlighting why and how for open ended questions, and then produces the final answer. While this method yields the most natural and fluent responses, it incurs higher latency. If a response is not available within 6 seconds, $A_3$ falls back on alternative pathways.

The second preferred strategy involves summarizing factual outputs from a data synthesizer $S_i$ by the LLM, contingent on semantic similarity between the synthesized facts and the original query, to derive a factual answer. A cosine similarity score is computed, and summarization is attempted only if the similarity exceeds predefined thresholds as shown in Table 4.

When the LLM summarization is used, lower similarity thresholds are employed to increase recall. In contrast, when LLMs are bypassed, higher thresholds are used to enhance precision. If the summarization process also exceeds 6 seconds, the fallback response is a raw fact from $S_i$ with a prefixed disclaimer indicating reduced confidence or precision. In this case, the most semantically similar fact measured by cosine similarity is selected.

4. Judge Agent, $A_4$

   If the response originates from LLM processing rather than direct extraction from $S_i$, it is passed to a judge agent for evaluation. The judge assesses each response along two dimensions where 100 is the highest and best score:
   - Factualness [0-100]: Measures the fidelity to underlying facts. Deductions in score are made for altered figures, fabricated statistics, contradictions, or misidentified player names.
   - Relevance [0-100]: Quantifies alignment with user intent, based on semantic overlap and topical focus.

   The Agent generates explanations for both dimensions such that the reasoning can be utilized within $MC$. A minimum threshold of 80 is required on both dimensions for the GenAI content to pass. If either threshold is not met, the most relevant factual candidate is rerouted to the final corrective agent.

5. Corrective Agent, $A_5$

   The final agent applies post-processing corrections using GenComp in the form of rules and regular expressions. The agent addresses formatting inconsistencies such as number representation, pluralization, round name, etc. Based on error analysis tailored to the selected LLM, additional adjustments such as predicate reordering may be performed to enhance fluency. Rationales for modifications are logged alongside judge evaluations and made available for transparency and interpretability within the $MC$ system.

The multi-agent architecture is explicitly designed for scalability and fault tolerance. Upon initialization, each $CE$ replica preloads 100 agent graphs into memory. These graphs serve as an isolated workflow, with one assigned to each incoming request, enabling each $CE$ instance to process up to 100 concurrent operations independently. In scenarios where capacity limits are reached or the system determines that a question cannot be confidently answered, a user-friendly fallback message is returned to the $MW$ application. This design ensures robust handling of high user traffic, knowledge base limitations, and out-of-domain queries ensuring system responsiveness under different operational loads.

## 6 Results

The development and deployment of a large-scale, real-time Match Chat, $MC$, assistant required interdisciplinary expertise spanning engineering, data science, and human-computer interaction design. To evaluate the system in a live setting, two practical user-centered studies were conducted with clearly defined participant segments.

A total of 33 users were selected to test the $MC$ feature during both Grand Slams. Participants were drawn from a predefined set of user personas with varying levels of engagement with tennis and digital sports products:
- Tennis Lover (58%): Highly engaged fans who follow tennis throughout the year.
- Sports Fan (9%): General sports enthusiasts with occasional tennis interest.
- Social Fan (3%): Users who engage primarily during high-profile matches and value engaging commentary.
- Two-Week Fan (15%): Casual viewers who tune in primarily for major tournaments.
- Passive Fan (15%): Users who engage sporadically, typically during major moments, and prefer summarized updates.

Each participant completed a structured 30-minute usability session designed to capture qualitative and quantitative feedback. Results from these sessions indicated that 81% of users found Match Chat helpful for contextualizing ongoing matches. Additionally, 69% of users reported the interface to be low-friction and easy to navigate, particularly for making tennis-related selections. 63% described the overall design as intuitive. However, 19% of users expressed uncertainty regarding the source of the information provided and 38% indicated a lack of understanding about how match predictions were generated. These findings highlight the importance of improving system transparency and explainability to build user trust and comprehension.

In terms of deployment at scale, Match Chat was used by nearly 1 million unique visitors across Wimbledon and the US Open. 65% of user interactions occurred during live matches, suggesting a strong preference for accessing real-time match statistics and contextual insights during tennis play. Overall, user feedback and engagement metrics indicate that the Match Chat assistant achieved high usability and performance in real-world, high-traffic environments.

### 6.1 Match Chat Speed

To evaluate the performance and responsiveness of the Match Chat system, a set of 544 gold-standard questions was created by two independent human annotators. These annotators used the structured data feeds outlined in Section 3 to generate reference responses, referred to as gold answers. In addition to measuring response quality shown in section 6.2, runtime performance metrics were collected to assess system efficiency. The overall responsiveness of Match Chat was found to be a key factor in maintaining user engagement.

As shown in Table 2, under small load conditions of up to 8 RPS, the agentic workflow, described in Section 5.4, exhibited acceptable response times that were, on average, close to or under 6 seconds. However, because the preferred GenAI pipeline involved injecting structured context from an agentic tool directly into an LLM, this configuration resulted in increased latency. To better understand system performance under real-world, high-traffic scenarios, stress tests were conducted. In one configuration involving 600 concurrent users at a sustained rate of 490 RPS without the GenAI Shield described in Section 5.1, the average response time reached 20.10 seconds. To mitigate this latency, the LLM infrastructure was scaled horizontally to handle 120 RPS, distributed across 4 geographic regions with the activation of the GenAI shield. In this way, every user-submitted question received a response in 6 seconds or less.

| Units are in seconds | Tool to LLM | Synthesizer to LLM | Synthesizer |
|---|---|---|---|
| Average Runtime | 6.42 | 1.24 | 0.21 |
| Standard Deviation | 3.02 | 0.76 | 0.09 |
| Maximum Time | 25.42 | 11.38 | 1.22 |

Table 2: Time statistics based on response times from Match Chat

A critical factor influencing latency was the length of tokens in both user prompts and LLM-generated outputs. Longer, more verbose prompts require additional encoding and decoding steps during inference, which significantly impacts response time. As demonstrated in Table 3, the average token length for common tennis-related questions from the gold set processed by the LLaMA 3-3 70B Instruct model was analyzed and correlated with observed latency. These results underscore the importance of token efficiency in optimizing real-time system performance. As a result, the maximum number of input characters allowable by a user was 100 characters. Through prompting, the length of the response was shortened with an average length of 34.97 tokens and generally a singular sentence.

| | Prompt Tokens | Completion Tokens | Total Tokens |
|---|---|---|---|
| Average Length | 1226.65 | 34.97 | 1261.61 |
| Standard Deviation | 31.73 | 16.25 | 37.21 |
| Maximum Token Length | 1327 | 207 | 1450 |

Table 3: Token length statistics based on Tool to LLM within Match Chat

## 6.2 Match Chat Accuracy

The same gold-standard questions were used to evaluate the accuracy of Match Chat. After a gold question was sent to Match Chat, the response was compared to the human annotator created gold answer. To evaluate the correctness of the agentic response, an automated AI-based evaluator or AI Judge was used to assess the factual accuracy and contextual relevance of the question, prompt, and answer tuples. This was used to verify the two judge dimensions of evaluation. After several experiments, a threshold of 0.8 across both judge dimensions were selected which resulted in a 92.83% judge pass rate. In other 7.17% of the time, the question was routed to the knowledge base, or a data synthetic answer was returned.

To help retrieve facts related to the user's question that helps in the cases when the data synthesizer is invoked, the predictive model described in section 5.3 classified each question. This classification was used within a series of experiments to set classification-based thresholds for each user question as shown in Table 4. A precision threshold was selected in the event that the data synthesizer was going to return a singular answer as Match Chat's response. If several facts from the data synthesizer were going to be summarized by an LLM, the recall threshold was used. Each threshold was a measure of cosine similarity between the answer and question embeddings.

| Category | Precision Threshold | Recall Threshold |
|---|---|---|
| Match Statistics | 0.69 | 0.4 |
| Player Statistics | 0.65 | 0.4 |
| Predictions type | 0.56 | 0.4 |
| Biographics type | 0.66 | 0.45 |
| Logistics type | 0.48 | 0.4 |
| Live Point-by-Point type | 0.80 | 0.4 |

Table 4: Factual thresholding from $S_i$.

The performance of the question classifier was critical to the routing, tool selection, and agentic processing within Match Chat. Over a training set of 1,263 questions tested on 116 samples, the Random Forest Classifier with 100 estimators achieved a precision of 84%, a recall of 85.7%, and an accuracy of 85.7%. The gap in error rate within practice was minimized by collecting clues through chat from the user and the responses of GenAI to automatically correct the question classifier based on GenComp rules. The following Figure 5 depicts the ROC curve for the classifier.

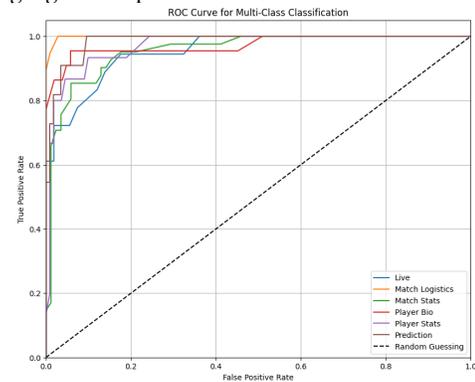

Figure 5: ROC of the question classifier.

To ensure content safety and appropriateness, the HAP pipeline was tested using an auxiliary database of 2,664 slurs, derogatory terms, and profanity, combined with standard Python-based derogatory term libraries. During ten user testing sessions, which collectively involved over 1,000 submitted questions, the system produced zero false negatives, indicating strong recall in identifying harmful inputs. However, a small number of false positives were observed, primarily involving misclassifications of player names. To address this, a curated acceptable list of terms, such as professional player first and last names, was incorporated. In a subsequent round of testing, the updated system achieved zero false positives, validating the effectiveness of the corrective measures.

From a performance standpoint, the Match Chat system demonstrated a strong balance between response time and answer accuracy. The GenAI Shield discussed in Section 5.1 routed queries to lightweight pipelines when latency exceeded six seconds, ensuring that system responsiveness remained within acceptable thresholds without decreasing output quality. This latency mitigation strategy helped maintain a seamless user experience even under high query volumes.

Match Chat's robust performance supported by high semantic accuracy, safe language generation, and adaptive response routing demonstrates the viability of deploying real-time, agentic systems in large-scale public settings. This work outlines a practical path forward for building performant AI assistants that prioritize speed, precision, and usability in live environments.

## ACKNOWLEDGMENTS

We would like to thank IBM Corporate Marketing and Project Support including Jeff Amsterdam, Miyuki Dalton, Brittany Caskey, Nick Wilkin, Eris Calhoun, Rhea Banerjee, Teddy Wellington, Monica Ellingson, Tyler Sidell and Kameryn Stanhouse for support. Gratitude goes to our property partners the USTA (US Open) and Wimbledon for enabling us to deliver live generative AI systems within their digital ecosystems.